\documentclass[letterpaper]{article}
\usepackage{aaai2026}
\usepackage{times}
\usepackage{helvet}
\usepackage{courier}
\usepackage[hyphens]{url}
\usepackage{graphicx}
\usepackage{graphicx}
\usepackage{natbib}
\usepackage{caption}
\usepackage{multirow}
\usepackage{makecell}
\usepackage{amsmath,amssymb,amsfonts}
\usepackage{mathrsfs}
\usepackage{appendix}
\usepackage{xcolor}
\usepackage{textcomp}
\usepackage{manyfoot}
\usepackage{booktabs}
\usepackage{algpseudocode}
\usepackage[ruled,linesnumbered]{algorithm2e}
\usepackage{listings}
\usepackage{enumitem}
\usepackage{cleveref}
\usepackage{subcaption}
\usepackage{booktabs}
\usepackage{placeins}

\usepackage{mdframed}
\usepackage{amsthm}
\usepackage{thmtools, thm-restate}
\usepackage{tikz-cd}

\usepackage{longtable}
\usepackage{calc}

\DeclareEmphSequence{}

\newcommand{\T}{^T}

\newcommand{\X}{\mathcal{X}}
\newcommand{\Y}{\mathcal{Y}}

\newcommand{\R}{\mathbb{R}}

\declaretheoremstyle[
    headfont=\bfseries, 
    bodyfont=\normalfont\itshape,
    headpunct={},
    spacebelow=\parsep,
    spaceabove=\parsep,
    mdframed={
        innertopmargin=6pt,
        innerbottommargin=6pt, 
        skipabove=\parsep, 
        skipbelow=\parsep} 
]{framedstyle}

\usepackage{newfloat}
\DeclareCaptionStyle{ruled}{labelfont=normalfont,labelsep=colon,strut=off}

\title{Oxytrees: Model Trees for Bipartite Learning}
\author{
    Pedro Ilídio\textsuperscript{\rm 1,\rm 2},
    Felipe Kenji Nakano\textsuperscript{\rm 1,\rm 2},
    Alireza Gharahighehi\textsuperscript{\rm 1,\rm 2},\\
    Robbe D'hondt\textsuperscript{\rm 1,\rm 2},
    Ricardo Cerri\textsuperscript{\rm 3},
    Celine Vens\textsuperscript{\rm 1,\rm 2}
}

\affiliations{
    \textsuperscript{\rm 1}
        KU Leuven, Campus KULAK,
        Dept. of Public Health and Primary Care
        Etienne Sabbelaan 53, Kortrijk, 8500, Belgium\\
        
    \textsuperscript{\rm 2}
        Itec, imec research group at KU Leuven,
        Etienne Sabbelaan 51, Kortrijk, 8500, Belgium\\
        
    \textsuperscript{\rm 3}
        Universidade de São Paulo,
        Instituto de Ciências Matemáticas e de Computação
        Av. Trab. São Carlense, 13566-590, São Carlos, São Paulo, Brazil\\
        
    \{%
    pedro.ilidio,
    felipekenji.nakano,
    alireza.gharahighehi,
    robbe.dhondt,
    celine.vens%
    \}@kuleuven.be, 
    cerri@icmc.usp.br
}

\begin{document}

\maketitle

\begin{abstract}
Bipartite learning is a machine learning task that aims to predict interactions between pairs of instances. It has been applied to various domains, including drug-target interactions, RNA-disease associations, and regulatory network inference. Despite being widely investigated, current methods still present drawbacks, as they are often designed for a specific application and thus do not generalize to other problems or present scalability issues.
To address these challenges, we propose Oxytrees: proxy-based biclustering model trees. Oxytrees compress the interaction matrix into row- and column-wise proxy matrices, significantly reducing training time without compromising predictive performance.
We also propose a new leaf-assignment algorithm that significantly reduces the time taken for prediction.
Finally, Oxytrees employ linear models using the Kronecker product kernel in their leaves, resulting in shallower trees and thus even faster training. Using 15 datasets, we compared the predictive performance of ensembles of Oxytrees with that of the current state-of-the-art.
We achieved up to 30-fold improvement in training times compared to state-of-the-art biclustering forests, while demonstrating competitive or superior performance in most evaluation settings, particularly in the inductive setting.
Finally, we provide an intuitive Python API to access all datasets, methods and evaluation measures used in this work, thus enabling reproducible research in this field.
\end{abstract}

\begin{links}
\link{Code}{https://github.com/pedroilidio/oxytrees2025}
\link{Python package}{https://github.com/pedroilidio/bipartite_learn}
\link{Extended version}{https://arxiv.org/}
\end{links}

\section{Introduction}
\label{sec:introduction}

Over the last two decades, technological advancements have led to the daily generation of enormous volumes of data, encompassing social media interactions, e-commerce transactions, IoT sensor outputs, and genomic sequences.
Such data are frequently presented in complex structures, which challenge the application of traditional machine learning algorithms.
In this work, we focus on the problem of bipartite learning, a machine learning task that involves predicting interactions within a network where two distinct types of instances are interconnected.
More specifically, bipartite learning is the task of modeling a function $(x_1,\; x_2) \mapsto y$ mapping a pair of objects $x_1$ and $x_2$ of different types to an output $y$ characterizing their interaction. Often, these datasets can be represented by two feature matrices, $X_1$ and $X_2$, which describe instances in each dimension, and an interaction matrix, $Y$, that holds their corresponding output~values.

Bipartite learning has been applied to \emph{several contexts}.
For instance, in bioinformatics, deep learning methods are employed for drug-target interaction \cite{huang_moltrans_2020,lin2023comprehensive, bagherian2021machine,liu2024ssldti}, miRNA-disease association \cite{tian2024mgcnss} and compound-protein interaction prediction \cite{tsubaki2019compound}.
Similarly, recommender systems can be seen as a user-item interaction problem \cite{aggarwal2016recommender}.

Regardless of the application, validating all possible interactions in $Y$ is unfeasible, e.g., it is impractical to verify all possible user-item or drug-target interactions. Consequently, many interactions are not directly observed but instead predicted using computational models trained on available data. It is often assumed that some of the training interactions labeled as negative are, in fact, unconfirmed positives, a setting commonly referred to as \emph{positive-unlabeled} (PU) learning~\cite{bekker2020learning}.
Many proposed methods can only predict interactions among instances present in the training set, i.e., they are not applicable to new instances~\cite{park_flaws_2012,schrynemackers_protocols_2013,pahikkala_toward_2015}, and hence, are not inductive~\cite{chapelle_semi-supervised_2006}. This limitation presents a significant challenge in the context of recommender systems, where it is referred to as the cold-start problem~\cite{gharahighehi_addressing_2022}. In the context of drug-protein interaction prediction, it is crucial to generalize to new disease proteins.
For that purpose, state-of-the-art techniques rely on deep learning to model the complex structure of the inputs (e.g., proteins as sequences, molecules as graphs)~\cite{huang_moltrans_2020,dehghan_tripletmultidti_2023}. As a result, they can only solve a specific type of interaction, severely compromising interpretability, and are computationally expensive, requiring large amounts of training data.

In this work, we investigate decision tree-based methods, due to their applicability to any type of interaction problem,
interpretability, capacity to handle problems with few training data, and state-of-the-art performance on tabular input~\cite{grinsztajn_why_2022,costa2023recent,blockeel2023decision}. Although promising, the current tree-based approaches to bipartite learning also pose significant scalability limitations. Furthermore, these decision trees fail to accurately represent sparse regions of the training space, since each leaf is a very simple model (the average label) of a small localized region.
In this paper, we propose a novel method, named \emph{Oxytrees}, as a solution to these limitations.

The core idea of Oxytrees is to aggregate the impurity of one of the dimensions of $Y$ into small matrices we call \textit{proxies}. The proxies are reused when evaluating multiple split candidates, aggregating only the remaining dimension when calculating the impurity of each candidate.
Additionally, Oxytrees introduce 
a \emph{faster leaf-assignment procedure}, for quicker inference of all interactions between large batches of instances; as well as
\emph{model trees for bipartite learning}.
\emph{Model trees}~\cite{quinlan1992learning,landwehr_logistic_2005,costa2023recent} return a (simple) function in the leaf nodes, rather than a constant.
This way, instance pairs arriving in the same leaf can still get very different predictions, allowing to more easily model complex relationships, and at the same time potentially allowing smaller tree sizes (with larger leaves\footnote{\textit{Oxytree} is the popular name of a genus of trees (\textit{Paulownia sp.}) that are not too tall, grow very fast, and have very large leaves.}), hence reducing the tree construction time even further.
Leaf models also enable trees to model functions that extend beyond the dense regions of the training set, where traditional decision trees struggle.

Two \emph{preliminary investigations} of our research groups relate to this study. \citet{alves2022twostep} use a single tree with gradient boosting models in the leaves with large label imbalance. The complex leaf models amplified the scalability issues.
We later~\cite{ilidio_fast_2024} introduced a preliminary version of the node-splitting procedure used by Oxytrees. In both studies, only a limited set of experiments was evaluated, and the performance improvements remained unclear.

For the present study, we built ensembles of Oxytrees and compared them to state-of-the-art techniques in bipartite learning. We selected methods applicable to a wide range of inductive interaction prediction tasks and compared them across 15 datasets.
Furthermore, we investigate the PU setting in detail by randomly masking 25\%, 50\%, and 75\% of the positive interactions.
To the best of our knowledge, this represents the most extensive comparative study to date in the context of inductive bipartite learning, which helps delineate the state-of-the-art of this emerging paradigm.
To assist future studies and allow reproducible research, we make all our algorithm implementations and preprocessed datasets \emph{publicly available},
using an accessible and intuitive Python API based on Scikit-Learn~\cite{pedregosa_scikit-learn_2011}.

\section{Formal Problem Definition}
\label{sec:background}

Consider two independent feature spaces $\X_1$ and $\X_2$, and an output space $\Y$.
We address the problem of modeling a function $f \colon \X_1 \times \X_2 \to \Y$ given only finite samples $X_1 \in \X_1^{n_1}$, $X_2 \in \X_2^{n_2}$, and $Y\in \Y^n$, with $Y$ representing known outputs of $f$ corresponding to dyads in $X_1 \times X_2$.
We refer to this paradigm as \emph{bipartite learning}.
We call $\X_1$ and $\X_2$ the two \emph{domains} or \emph{dimensions} of the problem.

In our specific learning setting, we focus on tasks with structured input with $\X_1 \subseteq \R^{m_1}$ and $\X_2 \subseteq \R^{m_2}$. Therefore, $X_1$ and $X_2$ can be represented as matrices $X_1=(X_1^{ij})$ and $X_2=(X_2^{ij})$. We use indices as superscripts and reserve the subscripts for descriptors of the matrix or vector as a whole.
We consider a single output, so $\Y \subseteq \R$ and an interaction matrix $Y$ can be built to hold the annotated interactions as $Y^{ij} = f(X_1^i,\; X_2^j)$.
We specifically focus on binary classification and therefore $\Y=\{0,\;1\}$. Furthermore, we assume the PU setting~\cite{bekker2020learning}, meaning that positive annotations are deemed confirmed, but negative annotations can be either ground-truth negatives or unannotated positives.

\section{Related work}
\label{sec:related work}

A handful of terms have been used in the literature to describe similar learning problems,
such as interaction prediction~\cite{schrynemackers_classifying_2015,pliakos_network_2019,chen_machine_2018,bagherian_machine_2020}, link prediction~\cite{lu_link_2011,zhou_progresses_2021}, dyadic prediction~\cite{menon_log-linear_2010,pahikkala_two-step_2014,jin_multitask_2017}, and network inference~\cite{park_flaws_2012,pahikkala_toward_2015,schrynemackers_protocols_2013,pliakos_network_2019}.
Link prediction, dyadic prediction, and network inference do not specify the existence of feature spaces ($X_1$ and $X_2$ can be derived from $Y$). Furthermore, they allow $X_1 = X_2$, not resulting in a bipartite network, as is also the case for interaction prediction.

To build inductive models for bipartite learning,
two common approaches reorganize the data to enable traditional algorithms. We call them \emph{data-based} adaptations, and they are classified as global and local~\cite{schrynemackers_classifying_2015}.
Global methods transform $X_1$ and $X_2$ into a single input space, whereas local methods build separate models for each domain~\cite{schrynemackers_classifying_2015}.

Unlike data-based approaches, some methods propose or modify machine learning estimators on a fundamental level to take maximum advantage of the bipartite learning setting.
We call them estimator-based adaptations.
\emph{Estimator-based} methods from the recent literature are mainly based on deep learning, matrix factorization or decision trees.
\emph{Deep learning} methods provide end-to-end solutions in which data are processed in their raw format. For example, graphs (compounds) and sequences (proteins), together with their known interactions, are directly used as input to the neural networks in drug-target bipartite learning.
\emph{Matrix factorization} methods focus on decomposing the interaction matrix in a lower-dimensional latent representation, which is then used to infer interactions~\cite{guo2024rise,chen2022review}. Neighborhood-Regularized Logistic Matrix Factorization (NRLMF) carries neighborhoods to the latent space to enable inductiveness~\cite{liu_neighborhood_2016,liu_lpi-nrlmf_2017,liu_predicting_2020}.
\emph{Decision tree-based methods}~\cite{costa2023recent,blockeel2023decision} are relatively new in bipartite learning. Predictive Bi-Clustering Trees (PBCT)~\cite{pliakos_global_2018} are an extension of Predictive Clustering Trees \cite{blockeel1998top} that allows splits using features from both $X_1$ and $X_2$.
In this way, a biclustering (i.e. two-dimensional partitioning) of the interaction matrix is obtained.
PBCT was then extended to ensembles~\cite{pliakos_network_2019}, and further combined with a pre-training step that imputes the unreliable negative annotations using NRLMF.
The resulting method was named BICTR, and showed state-of-the-art performance among the methods that are applicable to all bipartite learning problems~\cite{pliakos_drug-target_2020}.
Nonetheless, the method was only applied to the drug-target setting, and has scalability limitations both in induction and inference time.

\section{The Proposed Method: Oxytrees}
\label{sec:oxytrees}

This study introduces Oxytrees, proxy-based model trees for bipartite learning. It combines a refined training algorithm with a new batch inference procedure and leaf-specific models to generate outputs.

\subsection{Requirements for the Impurity Function}
\label{sec:oxytrees impurity}

Let $Y_\text{node}$ be a subset of
$Y$ reaching a given tree node.
Decision trees use an impurity function $I(Y_\text{node})$ to select a split in each tree node~\cite{breiman_classification_1984,blockeel1998top}.
Oxytrees require (see theorem C2) that $I(Y_\text{node})$ can be expressed in terms of two arbitrary functions $\mu$ and $\rho$:
\begin{equation}
    \label{eq:i constraint}
    \textstyle
    I_\text{Oxytrees} (Y_\text{node}) = \rho(\sum_{ij}\mu(Y_\text{node}^{ij}))
    \text{.}
\end{equation}
In this study, we specifically used the variance of the vectorized interaction matrices: $I_\text{Oxytrees} (Y_\text{node}) = \sigma^2(\text{vec}(Y_\text{node}))$. More explicitly:
\begin{equation}
    \label{eq:i bgso}
    I_\text{Oxytrees} (Y_\text{node})
    = \frac{\sum_{ij}(Y^{ij}_\text{node})^2}{n_\text{node}} - \left(\frac{\sum_{ij}Y^{ij}_\text{node}}{n_\text{node}}\right)^2
    \text{,}
\end{equation}
which corresponds to
\begin{align}
    \mu(z) = \begin{bmatrix} 1, & z, & z^2 \end{bmatrix}
    &&
    \rho(z_1, z_2, z_3) = \frac{z_3}{z_1} - \frac{z_2^2}{z_1^2}
    \text{.}
\end{align}

The current state-of-the-art BICTR~\cite{pliakos_drug-target_2020} considers columns or rows of $Y_\text{node}$ as multiple outputs, using $I(Y_\text{node})$ as the average of the variances of each column or row. This is incompatible with \cref{eq:i constraint} (see corollary C2.1) and therefore prohibits the proposed training optimization.

More intuitively, \cref{eq:i constraint} is equivalent to requiring $I(Y)$ to be invariant to the order in which dyads are presented during training (see theorem C1).
This is a standard assumption, the only exceptions being BICTR-related methods and local methods~\cite{schrynemackers_protocols_2013}.

\subsection{Using Proxies to Speed Up Training}
\label{sec:oxytrees training}

The impurity format in \cref{eq:i constraint} enables a significant improvement in computational performance.
Oxytrees start the split search in a node by building proxy matrices $\tilde Y_1$ and $\tilde Y_2$ that aggregate row- and column-wise impurities, respectively:
\begin{align}
    \label{eq:proxies general}
    \textstyle
    \tilde Y_1^i = \sum_{j} \mu(Y_\text{node}^{ij})
    &&
    \textstyle
    \tilde Y_2^j = \sum_{i} \mu(Y_\text{node}^{ij})
    \text{.}
\end{align}
Oxytrees then use $\tilde Y_1$ to evaluate all the horizontal split candidates in the node (that partition the rows of $Y_\text{node}$), and $\tilde Y_2$ to evaluate all the vertical split candidates in the node (that partition the columns).
Finally, the split $s(Y_\text{node}) = \{Y_\text{A},\; Y_\text{B}\}$ with the largest decrease in impurity $\Delta I$ is chosen:
\begin{equation}
    \label{eq:delta i}
    \textstyle
    \Delta I =
        \frac{1}{n}(
            n_\text{node}\;I(Y_\text{node})
            - n_\text{A}\;I(Y_\text{A})
            - n_\text{B}\;I(Y_\text{B})
        )
        \text{.}
\end{equation}

The reasoning behind the computational improvement is as follows. A large number of different candidate splits is usually evaluated in each node. For a given horizontal (resp. vertical) split, it is much more efficient to calculate $\Delta I$ from the precomputed $\tilde Y_1$ (resp. $\tilde Y_2$) than from the original $Y$, as this bypasses aggregation across columns (resp. rows).
As a consequence, the overall split search is faster using $\tilde Y_1$ and $\tilde Y_2$ for split evaluation, even if considering the extra time needed to build the proxy matrices.
Precisely, building an Oxytree has
the following computational complexity
\begin{equation}
    \label{eq:complexity oxytree}
    \text{BuildOxytree} \in \Theta(n_1^2 (\log (n_1) + m))
    \text{,}
\end{equation}
in which $m$ represents the number of features chosen to be evaluated at each node and $n_1 \propto n_2$ is the number of instances in each dimension. For PBCT we have
\begin{equation}
    \label{eq:complexity pbct}
    \text{BuildPBCT} \in \Theta(m n_1^2 \log (n_1))
    \text{.}
\end{equation}
As a result, Oxytrees reduce the training time by a factor of $\log (n_1)$  if $m \in \Omega(\log (n_1))$ and by a factor of $m$ otherwise, in comparison to PBCT-based methods (including the state-of-the-art, BICTR). A detailed derivation of \cref{eq:complexity oxytree} and \cref{eq:complexity pbct} is presented in
appendix D1,
while an empirical confirmation experiment is explored in \cref{sec:empirical complexity}.
Please refer to algorithms
FindBestSplitOxytree and FindBestSplitBICTR in appendix B
for pseudo-codes on the split search procedures.

\subsection{Efficient Batch Inference Procedure}
\label{sec:batch inference}

Consider a set of test row instances $X_\text{1, test}$ and a set of test column instances $X_\text{2, test}$.
The current state-of-the-art algorithm~\cite{pliakos_global_2018} traverses the tree once for each pair in $X_\text{1, test} \times X_\text{2, test}$ to make predictions (see \cref{fig:train test split}). Instead, our inference procedure receives the two entire feature matrices and, at each tree node, uses the split rule to partition either $X_\text{1, test}$ or $X_\text{2, test}$, depending on whether the rule represents a horizontal or vertical split. Each partition is passed to the corresponding child node, while the other, nonpartitioned, $X$ matrix is passed to both children. When the partitions $X_\text{1, leaf}$ and $X_\text{2, leaf}$ reach a leaf, we output the predictions for all the dyads in $X_\text{1, leaf} \times X_\text{2, leaf}$.
The algorithm is presented in further detail as
AssignLeavesOxytree
in appendix B.
Note that this procedure only evaluates each split rule once for each instance in the corresponding dimension, in contrast to the state-of-the-art BICTR, which performs redundant evaluations.
The resulting improvement in complexity is as follows.

Let $n_\text{test}$ and $n_\text{train}$ respectively be the numbers of test and training instances in each dimension.
The computational complexity of the Oxytrees batch inference procedure is given by
$\Theta(n_\text{test}^2)$ if $n_\text{test} \in \Omega(n_\text{train})$, and by $\Theta(n^2_\text{test} \log(n_\text{train}))$ otherwise.
For BICTR, the complexity is $\Theta(n^2_\text{test} \log(n_\text{train}))$ in all cases.
The most common scenario is $n_\text{test} \in \Theta(n_\text{train}) \subset \Omega(n_\text{train})$ (e.g. separating a fraction of the training data for testing, or performing k-fold cross validation). Hence, the batch inference procedure is expected to run $\log (n_\text{train})$ times faster in most situations, given sufficiently large $n_\text{train}$.
These results are derived thoroughly in
appendix D2
and empirically verified in \cref{sec:empirical complexity}.

\subsection{Generalized Output Function}
\label{sec:paulownias output function}

Accurate predictions require a careful trade-off between leaf sizes that are small enough to capture local patterns and large enough to generalize well.
Leaf models are a self-adaptive middle ground in this trade-off, allowing leaves to be larger and faster to train while maintaining representation power.
Oxytrees use Regularized Least Squares with the Kronecker product kernel (RLS-Kron) as the default leaf model, a bipartite adaptation of kernel Ridge regression proposed by \citet{van_laarhoven_gaussian_2011}.
We describe below how to obtain predictions with the RLS-Kron leaf models. An illustration is provided in
fig. A1.

Let $X_\text{1, leaf}$, $X_\text{2, leaf}$ and $Y_\text{leaf}$ denote the partition of the training set reaching a given leaf.
$x_\text{1, test} \in \X_1$ and $x_\text{2, test} \in \X_2$ are then test instances that were assigned to that leaf.
Given a context-specific similarity function $\phi_1: \X_1^2 \to \R$, we represent $x_\text{1, test}$ as a homonymous vector $\phi_\text{1, test}$ so that $\phi^i_\text{1, test} = \phi_1(x_\text{1, test},\; X^i_\text{1, leaf})$. 
Instances $x_\text{2, test}$ are represented analogously, and multiple individual representations are gathered in matrices $\Phi_\text{1, test}$ and $\Phi_\text{2, test}$.
Finally, the predicted outputs $\hat Y^{ij}$ between $\Phi^i_\text{1, test}$ and $\Phi^j_\text{2, test}$ are collectively obtained as
\begin{equation}
    \label{eq:rlskron output}
    \hat Y = \Phi_\text{2, test} \; W_\text{leaf} \;\Phi\T_\text{1, test}
    \text{.}
\end{equation}
The matrix $W_\text{leaf}$ is composed of learned linear coefficients and is determined by
\begin{equation}
    \label{eq:rlskron w}
    W_\text{leaf} = U_2
    \;[
        \Lambda
        \odot (U_2\T \; Y_\text{leaf} \; U_1)
    ]\;
    U_1\T
    \text{,}
\end{equation}
in which $U_1$ and $U_2$ are the matrices of eigenvectors of $\Phi_\text{1, leaf}$ and $\Phi_\text{2, leaf}$, respectively. $\Phi_\text{1, leaf}$ and $\Phi_\text{2, leaf}$, as suggested by their labels, are the symmetric similarity matrices corresponding to $X_\text{1, leaf}$ and $X_\text{2, leaf}$. The matrix $\Lambda$ has the same dimensionality as $Y_\text{leaf}$ and is defined in terms of the regularization parameter $\alpha$ and the vectors of eigenvalues $\lambda_1$ and $\lambda_2$ corresponding again to $\Phi_\text{1, leaf}$ and $\Phi_\text{2, leaf}$:
\begin{equation}
    \textstyle
    \Lambda^{ij} = ({\alpha + \mathbf{\lambda}_1^i \mathbf{\lambda}_2^{j}})^{-1}
    \text{.}
\end{equation}

\section{Experimental Setup}
\label{sec:experimental setup}

\subsection{Validation Procedure and Evaluation}
\label{sec:bipartite validation}

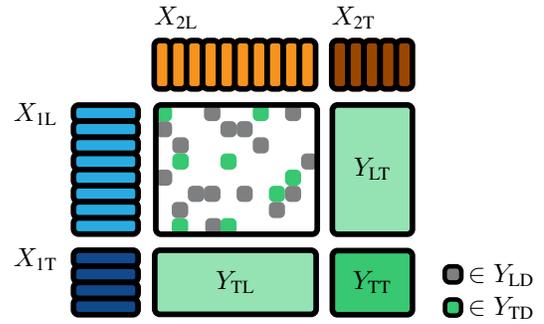
\begin{figure}[tbh]
    \def\svgscale{.3}
        \centering
        \begingroup%
    \makeatletter%
    \providecommand\color[2][]{%
        \errmessage{(Inkscape) Color is used for the text in Inkscape, but the package 'color.sty' is not loaded}%
        \renewcommand\color[2][]{}%
    }%
    \providecommand\transparent[1]{%
        \errmessage{(Inkscape) Transparency is used (non-zero) for the text in Inkscape, but the package 'transparent.sty' is not loaded}%
        \renewcommand\transparent[1]{}%
    }%
    \providecommand\rotatebox[2]{#2}%
    \newcommand*\fsize{\dimexpr\f@size pt\relax}%
    \newcommand*\lineheight[1]{\fontsize{\fsize}{#1\fsize}\selectfont}%
    \ifx\svgwidth\undefined%
        \setlength{\unitlength}{709.56848145bp}%
        \ifx\svgscale\undefined%
        \relax%
        \else%
        \setlength{\unitlength}{\unitlength * \real{\svgscale}}%
        \fi%
    \else%
        \setlength{\unitlength}{\svgwidth}%
    \fi%
    \global\let\svgwidth\undefined%
    \global\let\svgscale\undefined%
    \makeatother%
    \begin{picture}(1,0.60000009)%
        \lineheight{1}%
        \setlength\tabcolsep{0pt}%
        \put(0,0){\includegraphics[width=\unitlength,page=1]{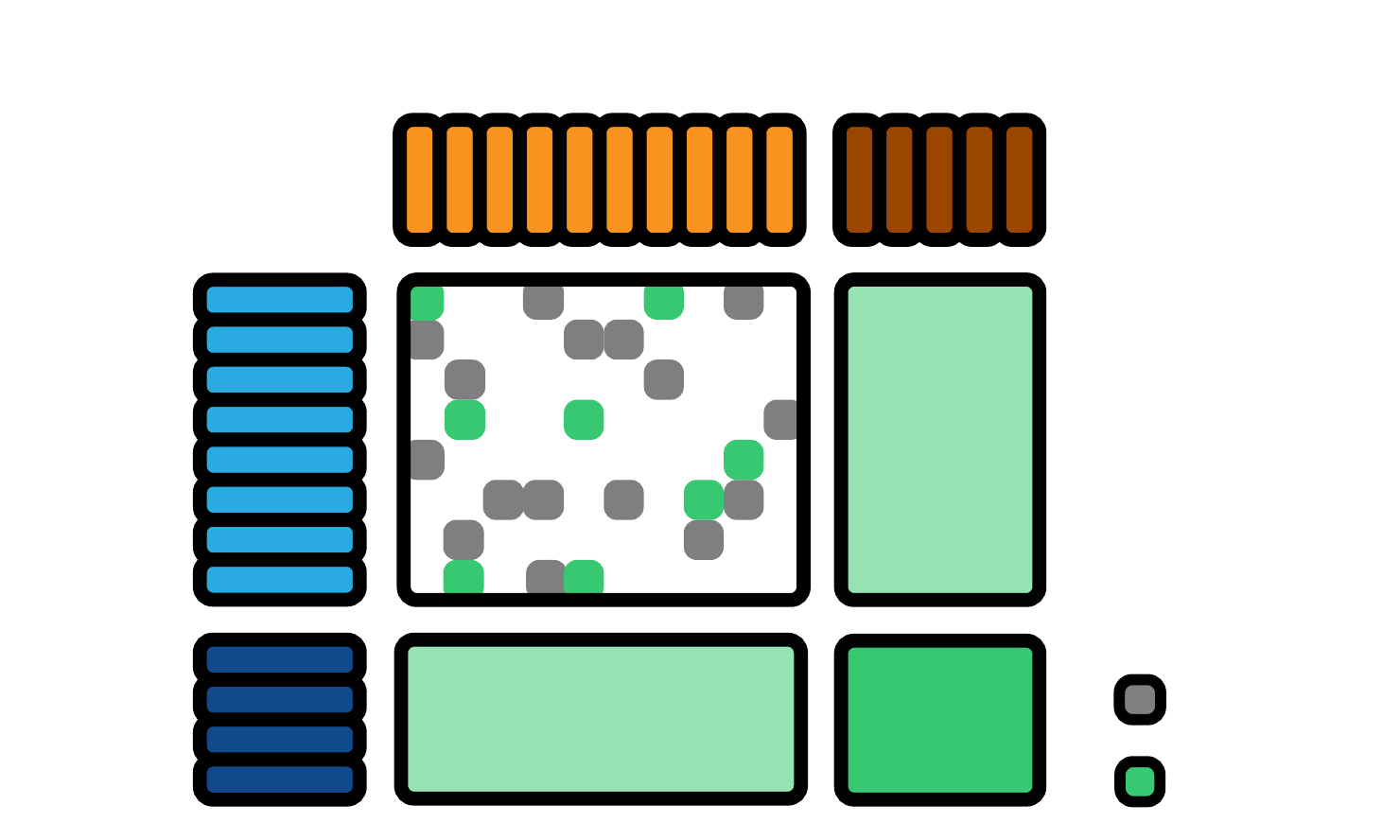}}%
        \put(0.1142856,0.37142863){\color[rgb]{0,0,0}\makebox(0,0)[rt]{\lineheight{1.25}\smash{\begin{tabular}[t]{r}$X_\text{1L}$\end{tabular}}}}%
        \put(0.28571412,0.54285719){\color[rgb]{0,0,0}\makebox(0,0)[lt]{\lineheight{1.25}\smash{\begin{tabular}[t]{l}$X_\text{2L}$\end{tabular}}}}%
        \put(0.1142856,0.11428577){\color[rgb]{0,0,0}\makebox(0,0)[rt]{\lineheight{1.25}\smash{\begin{tabular}[t]{r}$X_\text{1T}$\end{tabular}}}}%
        \put(0.59999984,0.54285719){\color[rgb]{0,0,0}\makebox(0,0)[lt]{\lineheight{1.25}\smash{\begin{tabular}[t]{l}$X_\text{2T}$\end{tabular}}}}%
        \put(0.67191852,0.27436032){\color[rgb]{0,0,0}\makebox(0,0)[t]{\lineheight{1.25}\smash{\begin{tabular}[t]{c}$Y_\text{LT}$\end{tabular}}}}%
        \put(0.84163196,0.08571434){\color[rgb]{0,0,0}\makebox(0,0)[lt]{\lineheight{1.25}\smash{\begin{tabular}[t]{l}$\in Y_\text{LD}$\end{tabular}}}}%
        \put(0.84163196,0.02857149){\color[rgb]{0,0,0}\makebox(0,0)[lt]{\lineheight{1.25}\smash{\begin{tabular}[t]{l}$\in Y_\text{TD}$\end{tabular}}}}%
        \put(0.42945104,0.07470452){\color[rgb]{0,0,0}\makebox(0,0)[t]{\lineheight{1.25}\smash{\begin{tabular}[t]{c}$Y_\text{TL}$\end{tabular}}}}%
        \put(0.67191852,0.07401614){\color[rgb]{0,0,0}\makebox(0,0)[t]{\lineheight{1.25}\smash{\begin{tabular}[t]{c}$Y_\text{TT}$\end{tabular}}}}%
    \end{picture}%
    \endgroup%

    \caption{Diagram of the test (T) and training (i.e. learning, L) sets in the bipartite context (see \cref{sec:bipartite validation}).}
    \label{fig:train test split}
\end{figure}

There are two ways of splitting bipartite datasets into \emph{training and test} sets to validate machine learning models: instance-wise and dyad-wise splits.
We analyze results for all split types by first performing an instance-wise split in each dimension, resulting in four partitions of the dataset, and then performing a dyad-wise split on the partition with training rows and training columns.
This results in the four test sets defined below and illustrated in \cref{fig:train test split}. L and T stand for learning (i.e., training) and test, respectively. D stands for dyads, referring to a dyad-wise split.
The naming scheme is inspired by \citet{schrynemackers_protocols_2013,pliakos_global_2018}.
\begin{itemize}
    \item \textbf{Learning set (LD):}
    $(X_\text{1L},\;X_\text{2L}, \;Y_\text{LD})$
    \item \textbf{Transductive test set (TD):}
    $(X_\text{1L},\;X_\text{2L}, \;Y_{\text{TD}})$
    \item \textbf{Semi-inductive test sets:}
    \begin{itemize}
        \item \textbf{Column-inductive test set (LT):}
            $
                (X_\text{1L},
                \;X_\text{2T},
                \;Y_\text{LT})
            $
        \item \textbf{Row-inductive test set (TL):}
            $
                (X_\text{1T},
                \;X_\text{2L},
                \;Y_\text{TL})
            $
    \end{itemize}
    \item \textbf{Inductive test set (TT):}
    $(
        X_\text{1T},
        \;X_\text{2T},
        \;Y_\text{TT}
    )$
\end{itemize}

For \emph{cross-validation}, each of the two domains was split into $k$ folds, using label stratification to encourage consistent fold densities. Each (row fold)-(column fold) pair can represent a TT set, yielding $k^2$ possible TT sets in total. The corresponding semi-inductive test sets were also collected for each TT set, and the training set and transductive test set were then generated.
The value of $k$ was adjusted for each learning task, using more folds for smaller datasets (\cref{tab:datasets}).

A percentage of the positive annotations was masked (replacing ones with zeros) to generate $Y_\text{TD}$.
We refer to this percentage as the \emph{positives masking percentage} (PMP).
All the missing annotations in the learning set (zeros that were not generated by masking) were also used as negative annotations for evaluation in the transductive setting. We explored four values for PMP: $0\%$, $25\%$, $50\%$, and $75\%$.

Regarding \emph{evaluation measures}, we have employed the area under the precision and recall curve (AUPRC) and the area under the ROC curve (AUROC), following  \citet{schrynemackers_protocols_2013,pliakos_global_2018,pliakos_drug-target_2020,liu_drug-target_2022}. 
Moreover, we have employed the Friedman test, followed by the post-hoc Nemenyi test, as suggested by \citet{demsar_statistical_2006}.
We report results for the semi-inductive settings altogether, averaging ranks of corresponding TL and LT scores.

\subsection{Datasets}
\label{sec:datasets}

\Cref{tab:datasets} describes the datasets used in this work.
As can be seen, we have employed a total of 15 datasets from several biological domains.
All features correspond to similarities measured between instances of the same input space.
Feature matrices in the test sets represent similarities with the training instances (avoiding data leakage).

\begin{table}[ht]
\centering
\small
\setlength{\tabcolsep}{1mm}
\begin{tabular}{llllrrl}
\toprule
Name & Domain & Size & Folds & Density \\
\midrule
DPI-N & Drug-NR & $26 \times 54$ & $4 \times 4$ & 6.4\% \\
TE-piRNA & TE-piRNA & $60 \times 93$ & $4 \times 4$ & 3.9\% \\
lnc-mi & LncRNA-miRNA & $44 \times 218$ & $3 \times 3$ & 10\% \\
DPI-G & Drug-GPCR & $95 \times 223$ & $3 \times 3$ & 3.0\% \\
Davis & Inhibitor-Kinase & $68 \times 442$ & $3 \times 3$ & 5.0\% \\
lnc-D & LncRNA-Disease & $156 \times 241$ & $3 \times 3$ & 6.6\% \\
lnc2Cancer & LncRNA-Cancer & $367 \times 106$ & $3 \times 3$ & 7.3\% \\
DPI-I & Drug-Ion channel & $204 \times 210$ & $2 \times 2$ & 3.5\% \\
MiRNA-D & MiRNA-Disease & $462 \times 252$ & $2 \times 2$ & 11\% \\
ERN & Gene-TF & $1164 \times 154$ & $2 \times 2$ & 1.8\% \\
SRN & Gene-TF & $1821 \times 113$ & $2 \times 2$ & 1.8\% \\
NPInter & LncRNA-Protein & $586 \times 446$ & $2 \times 2$ & 18\% \\
DPI-E & Drug-Enzyme & $664 \times 445$ & $2 \times 2$ & 1.0\% \\
KIBA & Inhibitor-Kinase & $2111 \times 229$ & $2 \times 2$ & 20\% \\
miRTarBase & MiRNA-Gene & $1873 \times 415$ & $2 \times 2$ & 7.1\% \\
\bottomrule
\end{tabular}
\caption{Summary of the datasets used in this study. Density refers to the percentage of positive annotations in each dataset. NR = nuclear receptor; TE = transposable element; TF = transcription factor.}
\label{tab:datasets}
\end{table}

DPI-N, DPI-G, DPI-I, DPI-E, SRN, ERN, Davis and KIBA are widely employed in the literature \cite{schrynemackers_classifying_2015,pliakos_network_2019,huang_moltrans_2020}. We have preprocessed NPInter and mirTarBase in our preliminary work~\cite{ilidio_fast_2024}. We have also collected five more datasets, TE-Pirna, LncRNA-miRNA, lncRNA-disease, lncRNA-cancer, and miRNA-disease, which, to the best of our knowledge, have never been used as benchmarks for interaction prediction in general.
Lnc2Cancer contains lncRNA-cancer interactions \cite{gao2021lnc2cancer}\footnote{http://bio-bigdata.hrbmu.edu.cn/lnc2cancer/}.
The datasets LncRNA-disease, LncRNA-miRNA and miRNA-disease were made available in \cite{sheng2023data}, where the authors collected data from several repositories.
TE-piRNA contains interaction between transposable elements (TEs) and piwi-RNAs (piRNAs)~\cite{dos2024transposable}.

\subsection{Comparison Methods}
\label{sec:baselines}

We present in \cref{tab:method-summary} a list of the comparison methods used in this work. We selected prominent methods from the literature that i) were able to infer interactions for new instances, and ii) were applicable across different bipartite learning problems. We provide more details about them in
appendix E.
As previously mentioned, the most common deep learning approaches do not meet criterion (ii). We thus include a baseline multi-layer perceptron (MLP)~\cite{hinton1990connectionist} under the data-based global adaptation~\cite{schrynemackers_protocols_2013} to still represent this model category. Following previous literature~\cite{huang_moltrans_2020,dehghan_tripletmultidti_2023}, we performed random undersampling of negative annotations, resulting in balanced training datasets for the MLP baseline.

\begin{table}[tbh]
\centering
\small
\setlength{\tabcolsep}{1mm}
\begin{tabular}{lll|cccc}
\toprule
\textbf{Method}
    & \textbf{Base technique}
    & \textbf{Adapt.}
    & \textbf{TF}
    & \textbf{Ex}
    & \textbf{EB}
    & \textbf{In}
    \\
\midrule
\textbf{MLP}
    & Deep learning
    & Global
    &
    &
    &
    & \checkmark
    \\
\textbf{RLS-avg}
    & Linear regression (LR)
    & Local
    &
    & \checkmark
    &
    &
    \\
\textbf{RLS-Kron}
    & LR + Kronecker product
    & ---
    &
    & \checkmark
    & \checkmark
    & \checkmark
    \\
\textbf{BLMNII}
    & LR + Nearest Neighbors
    & Local
    &
    & \checkmark
    &
    & \checkmark
    \\
\textbf{NRLMF}
    & Matrix factorization
    & ---
    &
    &
    & \checkmark
    & \checkmark
    \\
\textbf{WkNNIR}
    & Nearest-Neighbors
    & Local
    & \checkmark
    &
    & 
    & \checkmark
    \\
\textbf{BICTR}
    & Decision trees
    & ---
    &
    & \checkmark
    & \checkmark
    & \checkmark
    \\
\midrule
\textbf{Oxytrees}
    & Decision trees
    & ---
    & \checkmark
    & \checkmark
    & \checkmark
    & \checkmark
    \\
\bottomrule
\end{tabular}
\caption{%
    Summary of methods evaluated in our experiments. References:
    MLP~\cite{hinton1990connectionist}; 
    RLS-\{avg, Kron\}~\cite{van_laarhoven_gaussian_2011};
    BLMNII~\cite{mei2013drug};
    NRLMF~\cite{liu_neighborhood_2016};
    WkNNIR~\cite{liu_drug-target_2022};
    BICTR~\cite{pliakos_drug-target_2020}.
    See
    appendix E
    for hyperparameters of each method.
    Legend: Adapt. = bipartite adaptation strategy; TF = hyperparameter tuning-free; Ex = explainable; EB = Estimator-based adaptation (as opposed to data-based); In = inductive.
}
\label{tab:method-summary}
\end{table}

Our proposal does not enforce any particular ensembling technique.
Following BICTR~\cite{pliakos_drug-target_2020}, we used extremely randomized~\cite{geurts_extremely_2006} Oxytrees for this study. 

Regarding hyperparameter tuning, a nested 2 by 2 cross-validation procedure was utilized, using AUPRC to select the best hyperparameter combination for each outer fold.
The specific hyperparameter sets considered for each method are also presented in
appendix E.

\section{Results and Discussion}
\label{sec:results}

\subsection{Oxytrees Against Previous Proposals}
\label{sec:literature comparison}

\begin{figure*}[tbh]
    \includegraphics[width=.45\textwidth]{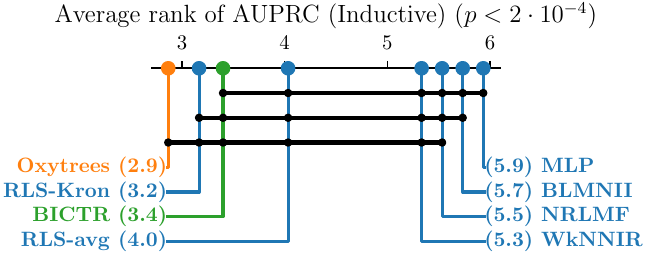}
    \hfill
    \includegraphics[width=.45\textwidth]{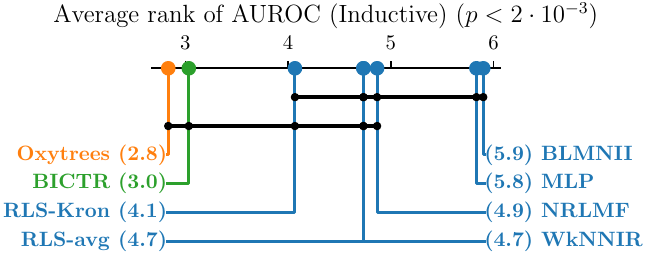}
    \caption{Comparisons of AUPRC and AUROC of the proposed Oxytrees against previous methods for bipartite learning. The inductive setting is analysed, for 0\% positives masking percent (PMP) (see \cref{sec:bipartite validation}).
    Further results in
    fig.~F1.
    }
    \label{fig:literature inductive subset}
\end{figure*}

Inductive AUPRC and AUROC results for 0\% PMP are presented in \cref{fig:literature inductive subset}. Similar results were found for other PMP values
(fig. F1).
Average values for AUROC and AUPRC in each dataset are presented by
table F2 and table F1,
respectively.
Oxytrees had the best average ranking for all but one of the inductive settings analyzed, and no statistically significant differences were found comparing Oxytrees to the state-of-the-art.
We also see that Oxytrees perform consistently better than RLS-Kron alone, indicating that the tree structure provides considerable improvement.
Figure~F2
presents results for the semi-inductive setting. Oxytrees, BICTR and RLS-Kron were the top performers, and the differences between their scores were not statistically significant in any case.
For the transductive setting
(fig.~F3),
NRLMF, BICTR and RLS-Kron had the best overall scores.

The scores in the inductive
(tables F1 and F2)
and semi-inductive
(tables F5-F8)
settings are considerably lower compared to the transductive setting
(tables F3 and F4),
which highlights the general difficulty of considering new instances.

On the other hand, NRLMF was one of the worst performers in the inductive setting compared to the other models. This shows that the original promising results of NRLMF in the transductive setting (found by \citet{liu_neighborhood_2016,liu_lpi-nrlmf_2017,liu_predicting_2020} and reproduced by us in
fig. F3).
do not translate into the inductive setting. This motivates the previously proposed combination of NRLMF and biclustering forests~\cite{pliakos_drug-target_2020}, restricting NRLMF to the transductive step of matrix completion and delegating the inductive modeling to the forest.
Furthermore, WkNNIR was not able to outperform BICTR in our experiments with 15 datasets, contrary to what was observed for drug-target interaction prediction~\cite{liu_drug-target_2022}.

To analyze the robustness of each model to missing positives, we compare their scores in different PMP settings relative to the setting with PMP = 0\%
(fig. F4).
As expected, the performances of all models decrease when we increase the PMP, with very few exceptions that we attribute to the intrinsic noise of these values.
BLMNII and RLS-avg were the models with the smaller performance drop in general, suggesting that the simplicity of local linear models makes them robust to missing labels.
The results for MLP were inconsistent, highlighting its sensitivity to the learning setting and hyperparameter choices.

\subsection{Ablation Study}
\label{sec:ablation}

\begin{figure}
    \includegraphics[width=0.49\linewidth]{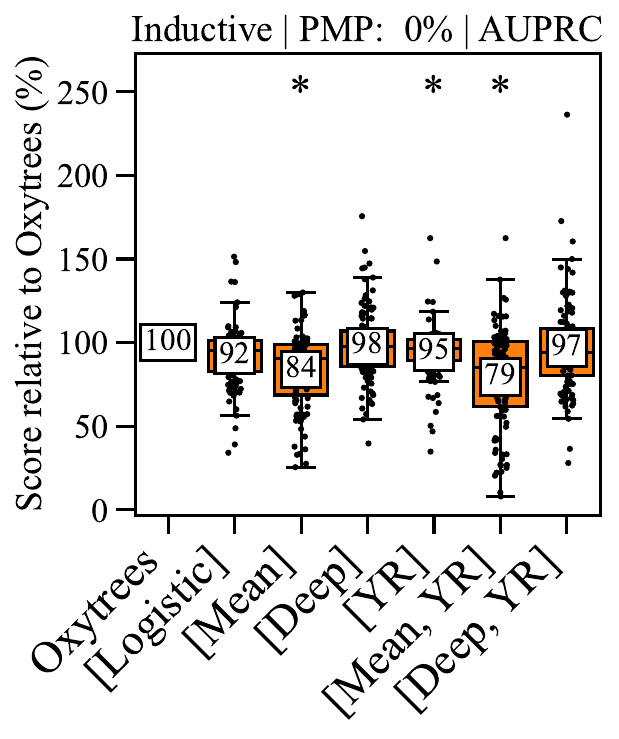}
    \hfill
    \includegraphics[width=0.49\linewidth]{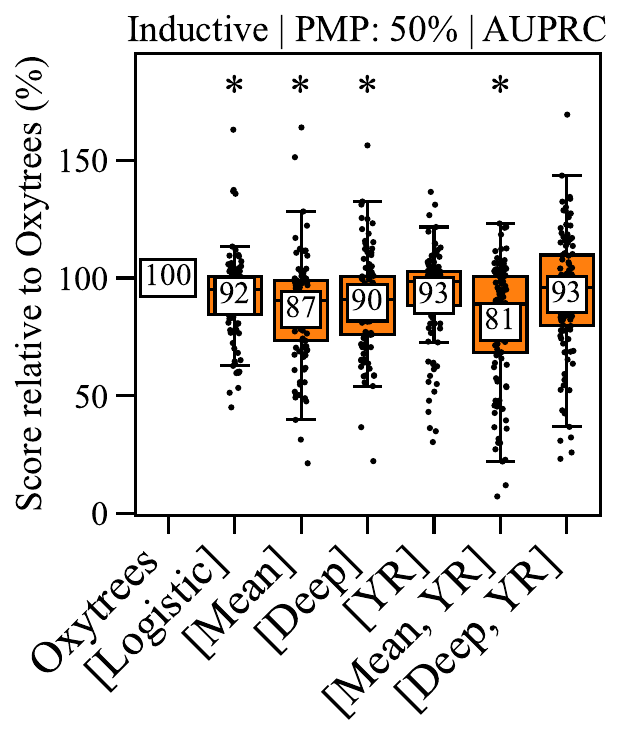}
    \caption{
    Performance comparison of ensembles of Oxytrees using different components (\cref{tab:ablation}). To compare across datasets, the scores are divided by the score of the main model (Oxytrees). Each point represents a cross-validation fold, and the mean value is presented in the white boxes.
    Asterisks indicate significance in comparison to Oxytrees ($p<0.05$, Wilcoxon signed-rank), averaging folds for each dataset.
    Remaining results are presented in
    fig. F6.
    }
    \label{fig:ablation}
\end{figure}

We compare ensembles of Oxytrees with and without the minimum leaf size constraint and different leaf models, as indicated in \cref{tab:ablation}. We also test the interaction matrix reconstruction (YR) technique proposed by \citet{pliakos_drug-target_2020}.
The results are presented in \cref{fig:ablation}.

Removing the leaf model (and using the average label as usual) significantly degraded the performance, even when using YR.
Using YR while enforcing minimum leaf dimensions of 5 by 5 was the worst overall strategy.
\begin{table}[htb]
\centering
\small
\setlength{\tabcolsep}{1mm}

\begin{tabular}{lccc}
\toprule
\textbf{Model}
    & \textbf{Leaf model}
    & \textbf{Min. 5 by 5 leaves}
    & \textbf{$Y$ rec.}
    \\
\midrule
\textbf{Oxytrees}
    & RLS-Kron
    & \checkmark
    &
    \\
\midrule
\textbf{[Mean]}
    & ---
    & \checkmark
    &
    \\
\textbf{[Logistic]}
    & Logistic reg.
    & \checkmark
    &
    \\
\textbf{[Deep]}
    & ---
    &
    &
    \\
\textbf{[YR]}
    & RLS-Kron
    & \checkmark
    & \checkmark
    \\
\textbf{[Mean, YR]}
    & ---
    & \checkmark
    & \checkmark
    \\
\textbf{[Deep, YR]}
    & ---
    &
    & \checkmark
    \\
\bottomrule
\end{tabular}
\caption{%
    Summary of method modifications compared in the ablation study.
    Interaction matrix reconstruction ($Y$ rec.) was performed with NRLMF~\cite{liu_neighborhood_2016}, following ~\citet{pliakos_drug-target_2020}.
    For the ``Deep" models, the trees were grown until all dyads in a node share the same label. For the ``Mean" models, the mean label was used for the prediction, as usual for decision trees.
}
\label{tab:ablation}
\end{table}

\subsection{Empirical Complexity Analysis}
\label{sec:empirical complexity}

\begin{figure}[tbh]
    \centering
        
    \includegraphics[width=0.49\linewidth]{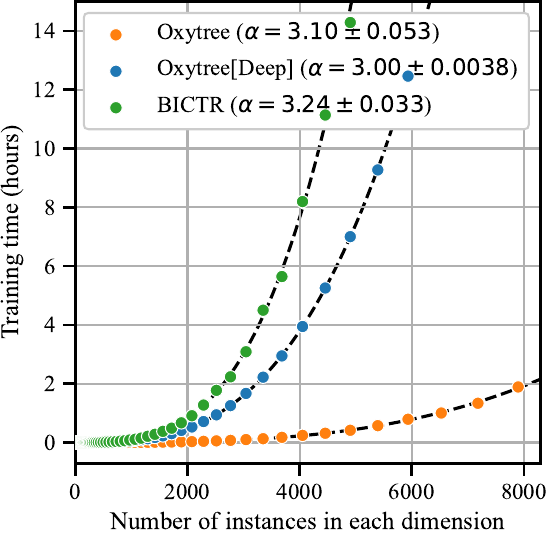}
    \includegraphics[width=0.49\linewidth]{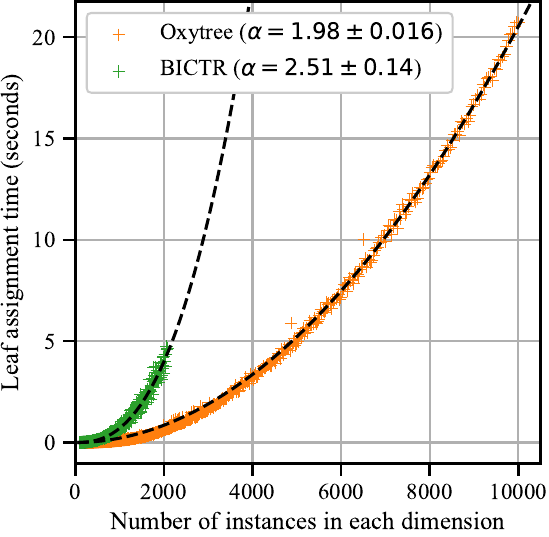}
    
    \caption{Empirical complexity analysis. Single trees were applied to artificial datasets of different dimensions (\cref{sec:empirical complexity}).
    We used the last 10\% of the points of each curve to approximate the asymptotic complexity as $\Theta (n^\alpha)$. $\alpha$ ($\pm$ standard dev.) is estimated as
    the slope of the linear regression in the log-log space.
    Further results in
    fig. F5.
    }
    \label{fig:empirical complexity}
\end{figure}

In this analysis, we compared the training and inference times of Oxytrees and BICTR, for different data sizes.
For each size $n_1$, we generated three $n_1$ by $n_1$ matrices with pseudo-random values to serve as $X_1$, $X_2$, and $Y$. We then built a single tree of the BICTR and Oxytrees ensembles for each $n_1$.
We also built an Oxytree until each leaf contained a single instance (Oxytree[Deep], \cref{tab:ablation}), to isolate the effect of the larger leaves.
To measure inference times, we built a completely random tree for each $n_1$ and used it twice to assign a leaf to each instance: through the BICTR procedure, and through the Oxytrees procedure (\cref{sec:batch inference}).
The results are shown in \cref{fig:empirical complexity}. The estimated complexities follow the theoretical expectations of \cref{sec:oxytrees training} (with $m = n_1$) and \cref{sec:batch inference}.
Plotting times of Oxytrees against BICTR (fig. F5), we measured that Oxytrees trained $35.49 \pm 0.75$ times faster and predicted $9.98 \pm 0.84$ times faster ($p < 4 \cdot 10^{-5}$, Wald).

\subsection{Dependency on Leaf Size}
\label{sec:leaf size}

\begin{figure}[tbh]
    \centering
    \includegraphics[width=\linewidth]{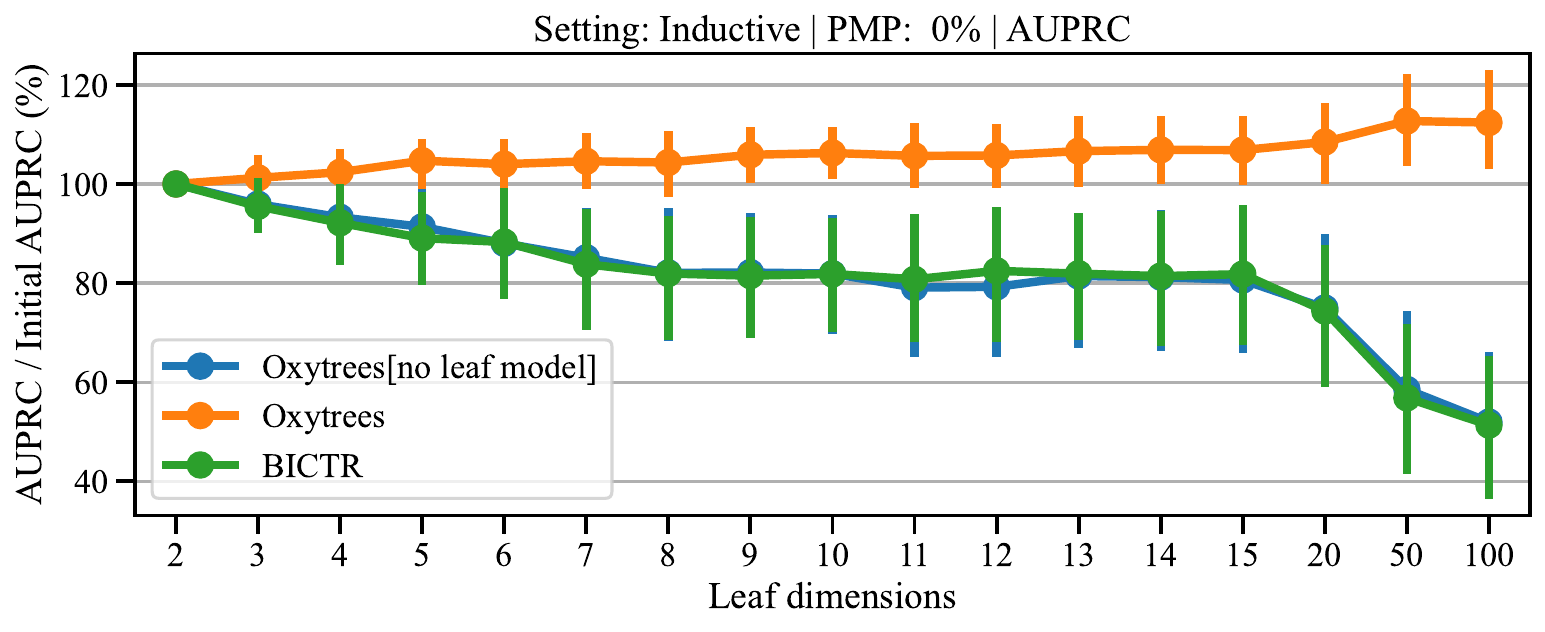}
    \includegraphics[width=\linewidth]{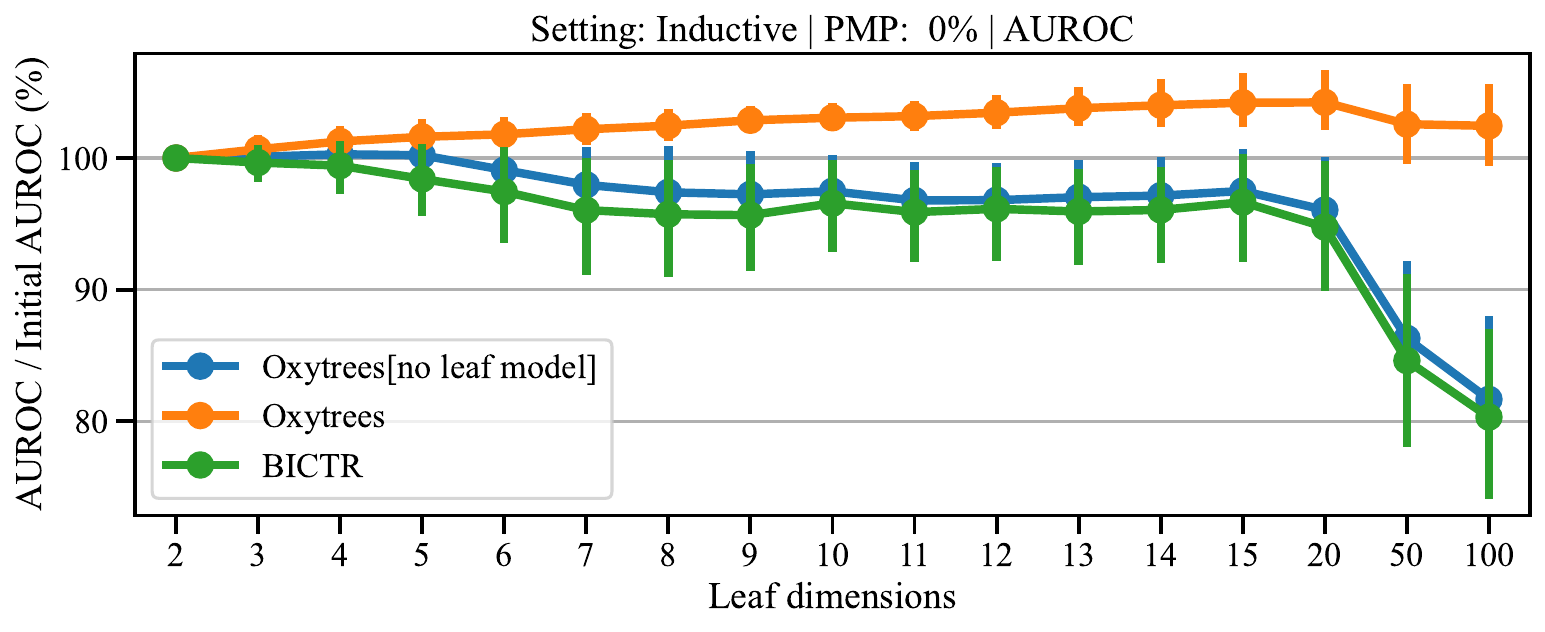}
    \caption{
    Performance of biclustering forests as a function of minimum leaf dimensions, relative to the initial performance with 2 by 2 leaves.
    Values in the $x$ axis represent the minimum number of instances in each dimension.
    Markers indicate mean and standard deviation over 15 datasets.
    }
    \label{fig:leaf size}
\end{figure}

We assess the effect of setting minimum leaf dimensions on the predictive performance of biclustering forests, comparing Oxytrees with RLS-Kron leaf models, Oxytrees without leaf models and BICTR. The results for the inductive setting are shown in \cref{fig:leaf size}.
Similar results were found for the other settings
(appendix F).
The performance of forests without leaf models drops when the leaves are enforced to be larger, especially for leaf dimensions larger than 20 by 20. On the other hand, Oxytrees with leaf models keep their performance stable even with larger leaves, with a slight score increase up until 20 by 20 leaves.
Therefore, adding RLS-Kron as leaf models enables shallower trees.

\subsection{Dependency on the Number of Trees}
\label{sec:number of trees}

\begin{figure}[ht!]
    \includegraphics[width=.48\linewidth]{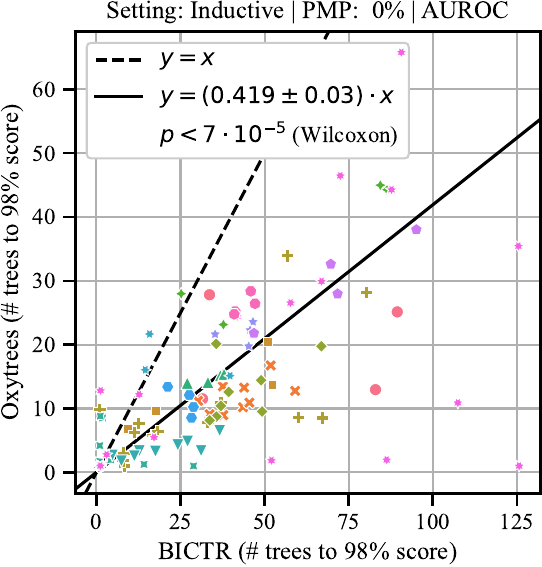}
    \hfill
    \includegraphics[width=.495\linewidth]{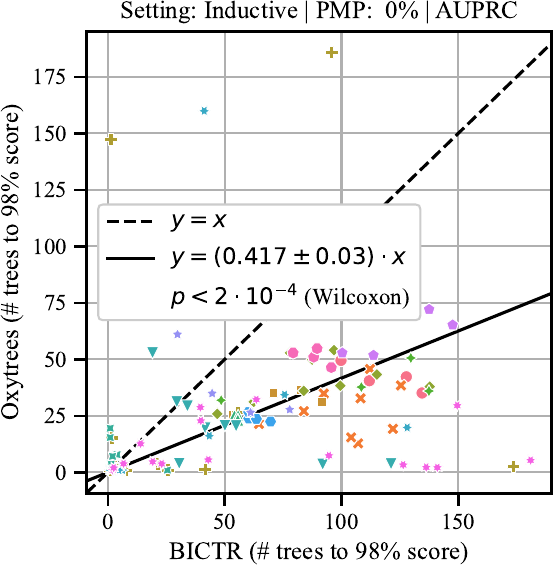}
    
    \caption{Comparison between Oxytrees and BICTR in terms of the expected number of trees to achieve 98\% of the score of 200 trees. Each fold is plotted, with different styles for each dataset.
     Remaining results can be found in
     fig. F8.
    }
    \label{fig:number of trees}
\end{figure}

We perform a bootstrapping procedure to evaluate how the performance of each biclustering forest varies with the number of trees.
For each forest, we started by building a total of 200 trees.
Then, we progressively added the outputs of each tree, calculating the resulting score in each step. We repeated the summing 50 times, with a random tree order each time.
From the plot of scores vs. number of trees, we used linear interpolation to estimate the number of trees required to achieve 98\% of the final score of 200 trees. The measurements for PMP 0\% are reported in \cref{fig:number of trees}.
Oxytrees required significantly fewer trees ($42\%\pm 3\%$) relative to BICTR ($p < 2 \cdot 10^{-4}$, Wilcoxon).

\section{Conclusion and Future Work}
\label{sec:conclusion}

We proposed an ensemble of proxy-based biclustering model trees: Oxytrees.
Oxytrees had competitive results among state-of-the-art techniques for inductive bipartite learning.
Against the state-of-the-art biclustering forest, BICTR, Oxytrees were $35.49 \pm 0.75$ times faster to train ($p < 4 \cdot 10^{-8}$) and predicted $9.98 \pm 0.84$ times faster ($p < 4 \cdot 10^{-5}$). They also required up to 50\% less trees to achieve 98\% of the performance of 200 trees ($p < 2 \cdot 10^{-4}$).
The model comparisons were mostly consistent across different masking percentages of positive annotations.
We provide datasets and efficient implementations of the algorithms compared in this study, in an accessible API based on Scikit-Learn.
As future work, we would like to explore the potential of Oxytrees for regressive bipartite learning, such as drug-target affinity prediction, and compare our results with deep learning techniques that work on the raw unstructured data (such as protein sequences and molecular structures).

\FloatBarrier

\section*{Acknowledgements}
The work received funding from the Flemish Government (AI Research Program), VOEWIAI02.
The authors thank FWO, 11A7U26N, 1235924N and 1S38025N.
R.C. thanks the Brazilian National Council for Scientific and Technological Development (CNPq), grant number 300934/2025-4, and the São Paulo Research Foundation (FAPESP), Brasil, Process Number \#2022/02981-8.

\bibliography{bibliography}
\vfill

\end{document}